\pdfoutput=1

\documentclass[11pt]{article}

\usepackage[preprint]{acl}

\usepackage{times}
\usepackage{latexsym}

\usepackage[T1]{fontenc}
\usepackage[utf8]{inputenc}
\usepackage{microtype}

\usepackage{amsmath}
\usepackage{booktabs}
\usepackage{enumitem}
\usepackage{inconsolata}
\usepackage{graphicx}
\usepackage{makecell}
\usepackage{subfigure}
\usepackage{tabularx}
\usepackage{multirow}

\title{PediaBench: A Comprehensive Chinese Pediatric Dataset for Benchmarking Large Language Models}

\usepackage{times}
\usepackage{latexsym}

\usepackage[T1]{fontenc}
\usepackage[utf8]{inputenc}
\usepackage{microtype}

\usepackage{amsmath}
\usepackage{booktabs}
\usepackage{enumitem}
\usepackage{inconsolata}
\usepackage{graphicx}
\usepackage{makecell}
\usepackage{subfigure}
\usepackage{tabularx}
\usepackage{multirow}

\title{PediaBench: A Comprehensive Chinese Pediatric Dataset for Benchmarking Large Language Models}

\author{
  \textbf{Qian Zhang\textsuperscript{1,2}},
  \textbf{Panfeng Chen\textsuperscript{1,2}},
  \textbf{Jiali Li\textsuperscript{1,2}},
  \textbf{Linkun Feng\textsuperscript{1,2}},
  \textbf{Shuyu Liu\textsuperscript{1,2}},
  \\
  \textbf{Heng Zhao\textsuperscript{3}},
  \textbf{Mei Chen\textsuperscript{1,2}},
  \textbf{Hui Li\textsuperscript{1,2,*}},
  \textbf{Yanhao Wang\textsuperscript{4,*}}
  \\
  \textsuperscript{1}State Key Laboratory of Public Big Data, College of Computer Science and Technology,\\Guizhou University, Guiyang, China
  \\
  \textsuperscript{2}Guizhou Engineering Laboratory for Advanced Computing and Medical Information\\Services, Guiyang, China
  \\
  \textsuperscript{3}College of Big Data and Internet, Shenzhen Technology University, Shenzhen, China
  \\
  \textsuperscript{4}School of Data Science and Engineering, East China Normal University, Shanghai, China
  \\
  \small{
    \textbf{\textsuperscript{*}Correspondence:} \href{mailto:cse.huili@gzu.edu.cn}{cse.huili@gzu.edu.cn}; \href{mailto:yhwang@dase.ecnu.edu.cn}{yhwang@dase.ecnu.edu.cn}
  }
}

\begin{document}

\maketitle

\begin{abstract}
The emergence of Large Language Models (LLMs) in the medical domain has stressed a compelling need for standard datasets to evaluate their question-answering (QA) performance.
Although there have been several benchmark datasets for medical QA, they either cover common knowledge across different departments or are specific to another department rather than pediatrics.
Moreover, some of them are limited to objective questions and do not measure the generation capacity of LLMs.
Therefore, they cannot comprehensively assess the QA ability of LLMs in pediatrics.
To fill this gap, we construct PediaBench, the first Chinese pediatric dataset for LLM evaluation.
Specifically, it contains 4,117 objective questions and 1,632 subjective questions spanning 12 pediatric disease groups.
It adopts an integrated scoring criterion based on different difficulty levels to thoroughly assess the proficiency of an LLM in instruction following, knowledge understanding, clinical case analysis, etc.
Finally, we validate the effectiveness of PediaBench with extensive experiments on 20 open-source and commercial LLMs.
Through an in-depth analysis of experimental results, we offer insights into the ability of LLMs to answer pediatric questions in the Chinese context, highlighting their limitations for further improvements.
Our code and data are published anonymously at \url{https://github.com/ACMISLab/PediaBench}.
\end{abstract}

\section{Introduction}

\begin{figure*}[t]
  \centering
  \includegraphics[width=\linewidth]{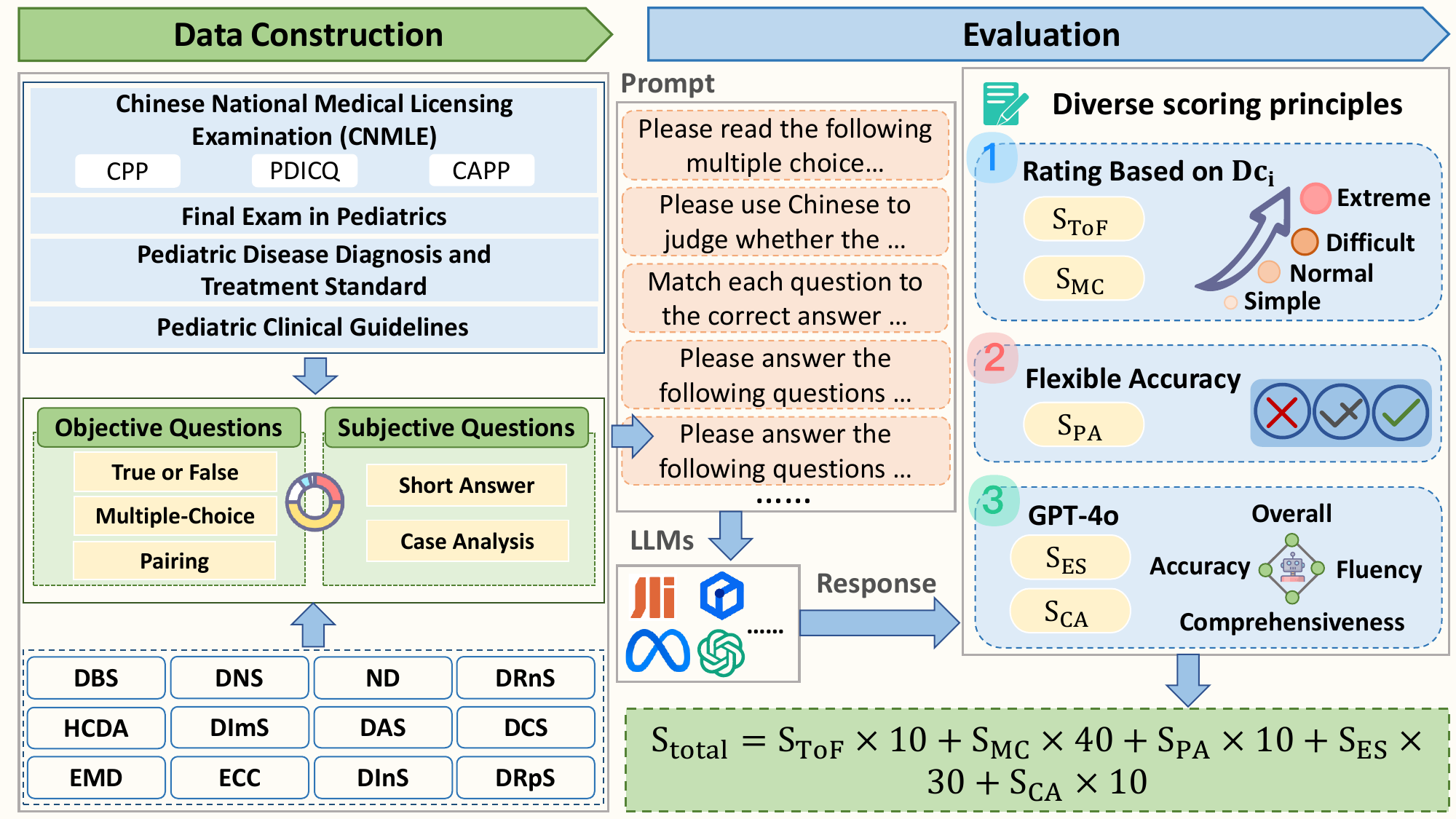}
  \caption{Illustration of the overall framework of PediaBench.}
  \label{figure1}
\end{figure*}

Question Answering (QA) is an important task in Natural Language Processing (NLP) that has received considerable attention over several decades \citep{10.1145/363707.363732, HIRSCHMAN_GAIZAUSKAS_2001, multi-hop-QA, Li_Zhou_Dou_2024}.
Recently, Large Language Models (LLMs) \citep{Baichuan2,qwen,gpt4}, with their remarkable language understanding, reasoning, and generation capabilities, have shown exceptional performance in QA tasks compared to classic deep learning methods \cite{bert}, especially for subjective questions.
Naturally, there is also an increasing interest in applying LLMs to medical QA \citep{bianque, singhal2023large, pulse, lievin2024can}.
In this paper, we focus on the QA tasks in \emph{pediatrics}, a medical department that involves the care of infants, children, adolescents, and young adults.
Since pediatrics often involves the manifestations and treatments of diseases that differ from those of adults, LLMs with common medical knowledge might not perform equally well on pediatric QA.
Therefore, evaluating the proficiency of LLMs in pediatric QA is an urgent need for their application in this domain.

Several medical QA benchmarks have been proposed in the literature \citep{PubMedQA, MedQA, MedMCQA, DDXPlus, CBLUE, Huatuo-26M, CMExam, CMB, pmlr-v219-wang23c, PromptCBLUE, MedBench, TCMBench}.
However, they still have some limitations that hinder their effectiveness in the pediatric context.
First, they are mostly general medical benchmarks across multiple departments and are not tailored for pediatrics.
As such, their coverage of knowledge in pediatrics is often very limited.
Second, most of them contain only objective questions, e.g., true/false and multiple-choice questions.
Although they can serve as an indicator of the capacity of LLMs to comprehend medical knowledge, they cannot assess the capacity of LLMs to generate medical texts.
In addition, some QA datasets originating from doctor-patient interactions in real-world scenarios \cite{Huatuo-26M} include some subjective questions to evaluate the conversation ability of LLMs.
However, they focus only on a limited number of common diseases and often lack depth of knowledge in medicine.
Therefore, existing benchmarks are insufficient to provide comprehensive evaluations of LLMs in terms of pediatric capability.

To address the above challenges, in this paper, we introduce PediaBench, the first comprehensive Chinese benchmark dataset in pediatrics.
As shown in Figure~\ref{figure1}, PediaBench contains 4,117 objective questions and 1,632 subjective questions collected from various sources, such as the Chinese National Medical Licensing Examination (CNMLE), final exams of universities in medicine, pediatric disease diagnosis and treatment standards, and clinical guidelines.
PediaBench encompasses five distinct types of questions, namely true or false, multiple choice, pairing, essay/short answer, and case analysis, across 12 typical pediatric disease groups.
Furthermore, to provide an accurate evaluation of the performance of each LLM for QA in pediatrics, we use a scoring criterion that combines difficulty levels and automatic scoring. On the one hand, we assign different scores to objective questions based on the difficulty factor; on the other hand, we adopt the Language-Model-as-an-Examiner (LME) framework \cite{LME} to score the objective questions based on GPT-4o.
Finally, we conduct extensive experiments with 20 LLMs on the PediaBench dataset and provide an in-depth analysis of the performance of LLMs for QA in pediatrics.
Our main contributions are summarized as follows:
\begin{itemize}
    \item We introduce PediaBench, a high-quality QA dataset specific to pediatrics in the Chinese context. We also devise an integrated scoring scheme to measure the QA performance of each LLM on PediaBench.
    \item We evaluate PediaBench with 20 LLMs, including open-source and commercial general-purpose models of different scales and specialized models in the medical domain. The results indicate that PediaBench is a challenging dataset that can gauge the capacities of LLMs in terms of pediatric QA.
    \item We provide a detailed analysis of the results, highlighting the limitations of current LLMs and suggesting their potential opportunities in this emerging domain.
\end{itemize}

\section{Related Work}

\paragraph{General LLM Benchmarks}
The rapid advances in LLMs \citep{LLaMA, Baichuan2, ChatGLM3-6B, qwen, gpt4} have underscored the need for benchmarks to evaluate their performance in a variety of NLP tasks.
To this end, a large number of benchmarks specifically designed for LLMs, e.g., \citep{GLUE, CLUE, SuperCLUE, gaokao, AGIEval, C-EVAL, CMMLU, BIG-Bench, XieZhi}, were proposed.
We refer interested readers to \citep{10.1145/3641289} for an extensive survey.
These general benchmarks often require LLMs to answer questions from standard examinations \citep{XieZhi, C-EVAL} in a broad spectrum of domains to evaluate their capacity in text understanding, logical reasoning, calculation, generation, etc.
However, although medical questions are included in some of these benchmarks, they are about common knowledge that can be answered without specialization, which cannot accurately reflect the proficiency of LLMs in medical tasks.

\paragraph{Medical LLMs and Benchmarks}
More recently, many efforts have been made to build specialized LLMs in the medical domain, e.g., DoctorGLM \citep{DoctorGLM}, ChatDoctor \citep{ChatDoctor}, BianQue \citep{bianque}, PMC-LLaMA \citep{PMC-LLaMA}, BioMistral \citep{BioMistral}, MEDITRON \citep{MEDITRON-70B}, ZhongJing \citep{zhongjing}, and QiLin-Med \citep{Qilin-Med}.
Generally, they use general-purpose LLMs as foundation models, construct training corpora with medical articles, textbooks, guidelines, etc., and fine-tune foundation models on the training corpora to inject medical knowledge.

With the development of medical LLMs, there has also been an increasing interest in benchmarking the medical knowledge of LLMs.
\citet{PubMedQA} constructed PubMedQA, a biomedical QA dataset from PubMed abstracts.
MedQA \citep{MedQA}, MedMCQA \citep{MedMCQA}, and CMExam \citep{CMExam} consisted of questions from standardized medical examinations.
The datasets above contain only multiple-choice questions for evaluation and cannot fully capture the generation capability of LLMs.
CMB \citep{CMB} and MedBench \citep{MedBench} incorporated clinical case analysis into the evaluation.
CBLUE \citep{CBLUE} and PromptCBLUE \citep{PromptCBLUE} included eight NLU tasks (e.g., named entity recognition, information extraction, and sentence classification) to evaluate the capabilities of models for language understanding in medicine.
\citet{DDXPlus} provided DDXPlus, a large-scale medical diagnosis dataset.
Huatuo-26M \citep{Huatuo-26M} contained a large number of real-world medical dialogues.
\citet{Rarebench} built a benchmark dataset with an emphasis on rare diseases.
\citet{TCMBench} collected a benchmark dataset that focuses on traditional Chinese medicine.
These datasets have contained subjective questions to evaluate the generation and conversation ability of LLMs.
However, all existing datasets have poor coverage of knowledge in pediatrics.
In addition, they usually use semantic similarity metrics \citep{10.3115/1073083.1073135,lin-2004-rouge} to evaluate subjective questions, which often leads to higher scores for longer answers and cannot fully reflect the proficiency of LLMs.

\begin{figure*}[t]
    \centering
    \includegraphics[width=\linewidth]{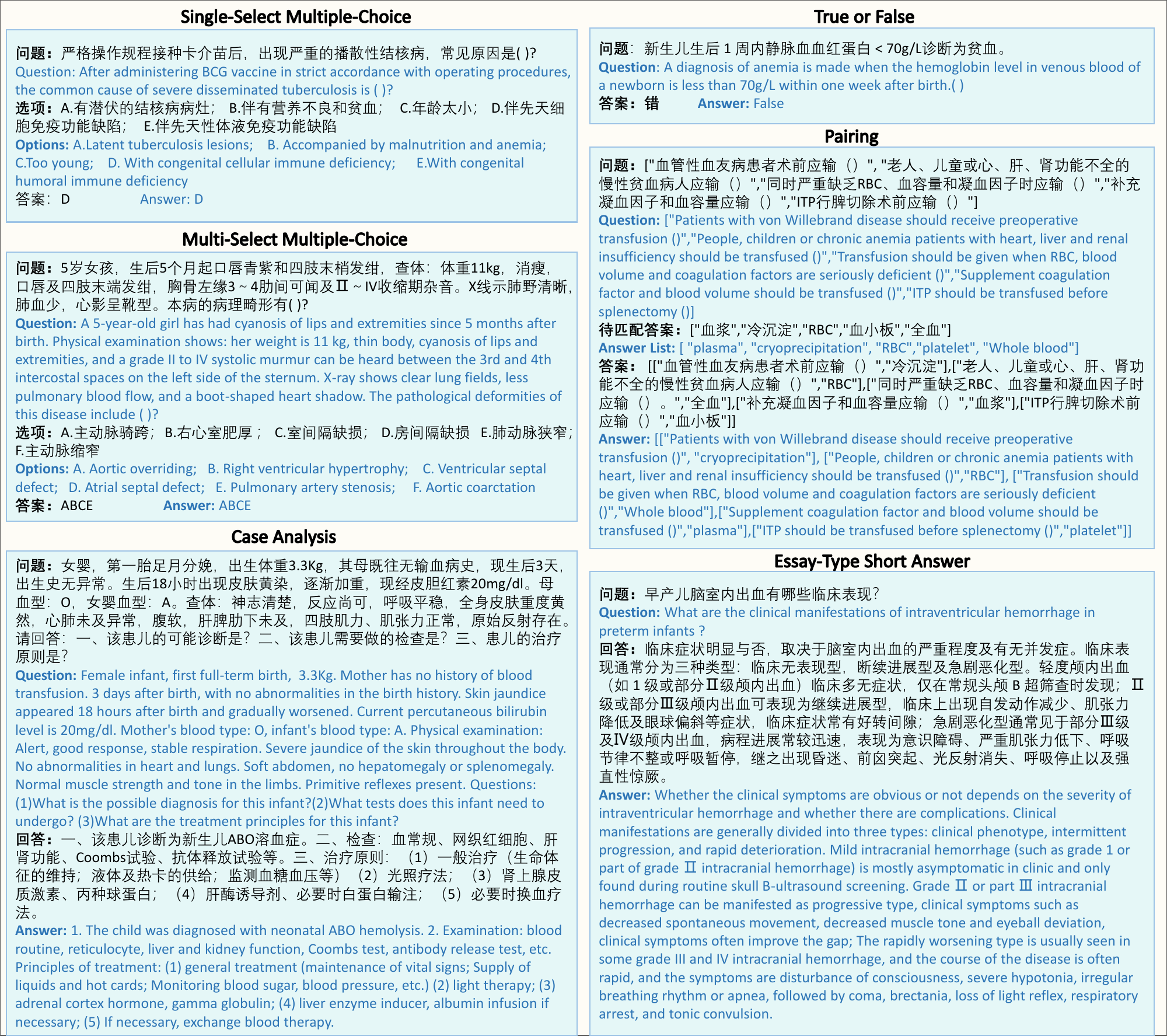}
    \caption{Examples for different types of questions and their answers in PediaBench.}
    \label{figure2}
\end{figure*}

\section{The PediaBench Dataset}
\label{sec-dataset}

This section provides a description of the construction process of the PediaBench dataset, as illustrated in Figure~\ref{figure1}.

\subsection{Question Types}

To assess how well an LLM can serve as an AI assistant for pediatricians, PediaBench incorporates the following five typical types of medical questions for evaluation:
\begin{itemize}
    \item \textbf{True or False (ToF):} This type of question asks whether a statement is factual. It requires an LLM to match the statement with its corresponding concepts and facts in the corpus, to understand their semantic meanings, and to reason about them so as to detect possible errors and contradictions.
    \item \textbf{Multiple Choice (MC):} This type of question asks for the selection of one (or more) appropriate choices from multiple candidates to complete a sentence or answer a question. It requires an LLM to distinguish between similar or related concepts. Some questions also evaluate the mathematical and logical skills of an LLM, as basic calculations are essential to obtain the correct answer.
    \item \textbf{Pairing (PA):} This type of question requires exactly matching all sentences with their corresponding missing words from the candidate list. Distinguishing among similar concepts is also essential for PA. However, since any mismatch leads to an entirely erroneous answer, PA is even more challenging than MC.
    \item \textbf{Essay/Short Answer (ES):} This type of question asks one to elaborate on a specific concept. It requires an LLM to generate coherent and accurate text relevant to the concept.
    \item \textbf{Case Analysis (CA):} This type of question presents an LLM with a description of a particular instance and asks the LLM to make a medical diagnosis and provide treatment measures.
    It can comprehensively evaluate the medical capacity of an LLM in terms of comprehension, reasoning, and problem-solving.
\end{itemize}
Figure~\ref{figure2} presents several examples of different types of questions and their sample answers in the PediaBench dataset.

\begin{table}[t]
    \centering
    \small
    \setlength{\tabcolsep}{4pt}
    \begin{tabular}{ccccc}
        \toprule
        \textbf{\makecell{Question\\Type}} &   \textbf{\#Questions} &   \textbf{\makecell{Data\\Source}} &   \textbf{\makecell{Disease\\Groups?}} & \textbf{Measure}\\
        \midrule
        ToF & 258 & \makecell{FE,\\PCG} & Yes & Accuracy\\
        \midrule
        MC & 3,576 & \makecell{CPP,\\CAPP,\\PDICQ} & Yes & Accuracy\\
        \midrule
        PA & 283 & \makecell{CPP,\\CAPP,\\PDICQ} & Yes & Accuracy\\
        \midrule
        ES & 1,565 & \makecell{FE,\\DTSPD,\\PCG} & Yes & LME\\
        \midrule
        CA & 67 & FE & No & LME\\
        \bottomrule
    \end{tabular}
    \caption{Statistics of the five question types in PediaBench. For \emph{data source}, ``FE'' stands for final exams of universities, ``DTSPD'' stands for diagnosis and treatment standards for pediatric diseases, ``PCG'' stands for pediatric clinical guidelines, ``CPP'' stands for Clinical Practicing Physician Examination, ``CAPP'' stands for Clinical Assistant Practicing Physician Examination , and ``PDICQ'' stands for Pediatric Doctor In-Charge Qualification Examination.}
    \label{table1}
\end{table}

\begin{table*}[t]
    \centering
    \small
    \begin{tabular}{ccccccc}
        \toprule
        \textbf{Disease Group} &   \textbf{Abbreviation} &   \textbf{MC} &   \textbf{ToF} &   \textbf{PA} &   \textbf{ES} &   \textbf{CA}\\
        \midrule
        Renal system &   DRnS &   293 &   21 &   24 &   139 &   \multirow{12}*{Unclassified (67)} \\
        Emergency and critical care &   ECC &   270 &   20 &   21 &   104 &   \\
        Infection system &   DInS &   297 &   20 &   20 &   150 &   \\
        Blood system &   DBS &   294 &   22 &   24 &   122 &   \\
        Cardiovascular system &   DCS &   295 &   21 &   20 &   138 &   \\
        Immune system &   DImS &   294 &   21 &   20 &   128 &   \\
        Respiratory system &   DRpS &   299 &   21 &   26 &   132 &   \\
        Endocrine or metabolic diseases&   EMD &   316 &   22 &   21 &   145 &   \\
        Health care and developmental Abnormalities &   HCDA &   318 &   26 &   40 &   149 &   \\
        Neonatal diseases &   ND &   293 &   22 &   21 &   144 &   \\
        Alimentary system &   DAS &   314 &   22 &   21 &   104 &   \\
        Nervous system &   DNS &   293 &   20 &   25 &   110 &   \\
        \bottomrule
    \end{tabular}
    \caption{Statistics of the number of questions of each type from the 12 disease groups in PediaBench.}
    \label{count-by-disease}
\end{table*}

\subsection{Data Collection and Processing}

The questions in PediaBench are collected from diverse yet reliable sources, including the Chinese National Medical Licensing Examination, final exams of universities in medicine, and pediatric disease diagnosis and treatment standards and clinical guidelines.
Next, we will describe each of them.

\paragraph{Chinese National Medical Licensing Examination (CNMLE)}
We have gathered 3,576 multiple-choice questions from CNMLE, including the Clinical Practicing Physician Examination (CPP), the Clinical Assistant Practicing Physician Examination (CAPP), and the Pediatric Doctor In-Charge Qualification Examination (PDICQ).
3,383 of them were from the question bank of PDICQ, and the rest were from the question banks of CPP and CAPP and relevant to pediatrics.
There are 50 questions with more than one correct answer.
For these questions, an answer is considered correct if it includes and only includes all correct options.

\paragraph{Final Exams in Medicine}
We have collected the final exams of universities in medicine from their official websites.
We manually extracted 258 true-or-false questions, 167 essay/short-answer questions, and 67 case analysis questions in pediatrics from the original PDF files.

\paragraph{Pediatric Disease Diagnosis and Treatment Standards \& Clinical Guidelines}
We also curated 1,398 essay/short-answer questions from the \emph{Pediatric Disease Diagnosis and Treatment Standards} series and \emph{Clinical Guidelines}, covering critical aspects such as etiology, diagnostic criteria, treatment plans, and preventive measures for diseases across diverse pediatric specialties.

\paragraph{Manual Construction}
We manually constructed a set of 283 pairing questions based on the multiple-choice questions with shared stems or answer options.
Since pairing questions require an exact matching between all missing parts in the stems and options, they are more challenging than multiple-choice questions.

\paragraph{Classification of Disease Groups}
The questions curated from the \emph{Pediatric Disease Diagnosis and Treatment Standards} series and \emph{Clinical Guidelines} have been annotated with disease groups to which they belong, referring to the International Classification of Diseases (ICD-11) standard issued by the WHO.
We further annotated the remaining unclassified questions. Specifically, we used GLM-4 \cite{GLM} to guide disease group classification.
We first wrote cue phrases for GLM-4 to suggest a disease group for each question.
Then, we manually eliminated duplicate questions, double-checked the classification results of GLM-4, and corrected the misclassified ones.

\subsection{Dataset Statistics}

As shown in Table~\ref{table1}, the PediaBench dataset consists of 5,749 questions, including 258 true-or-false questions, 3,576 multiple-choice questions, 283 pairing questions, 1,565 essay/short-answer questions, and 67 case analysis questions.
Except for case analysis questions, the remaining 5,682 questions are organized into 12 distinct disease groups, namely, diseases of the renal system, emergency and critical care, diseases of the infection system, diseases of the blood system, diseases of the cardiovascular system, diseases of the immune system, diseases of the respiratory system, endocrine or metabolic diseases, health care and developmental abnormalities, neonatal diseases, diseases of the alimentary system, and diseases of the nervous system.
Note that case analysis questions do not need to be classified because (1) they require an LLM to diagnose the disease groups to which the instance belongs according to symptoms and (2) the instance might belong to more than one disease group.
The number of questions for different disease groups in PediaBench is shown in Table~\ref{count-by-disease}.
We can see that PediaBench features a balanced distribution among all disease groups, comprehensively covering a wide spectrum of prevalent and rare pediatric diseases and establishing an unbiased benchmark.

\subsection{Evaluation Criteria}

\begin{figure}[t]
    \centering
    \subfigcapskip=0pt
    \subfigure[ToF]{\includegraphics[width=0.475\linewidth]{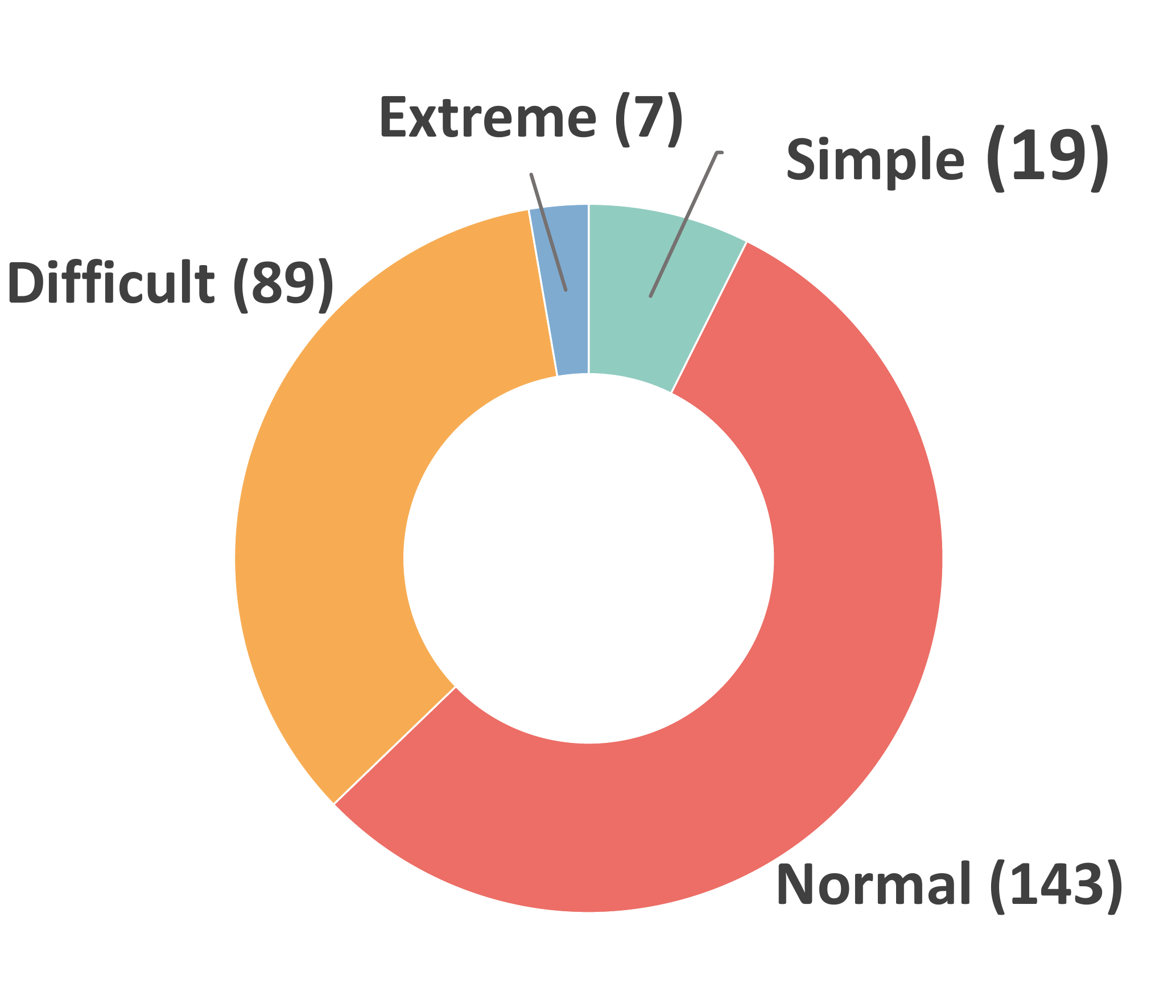}}
    \subfigure[MC]{\includegraphics[width=0.475\linewidth]{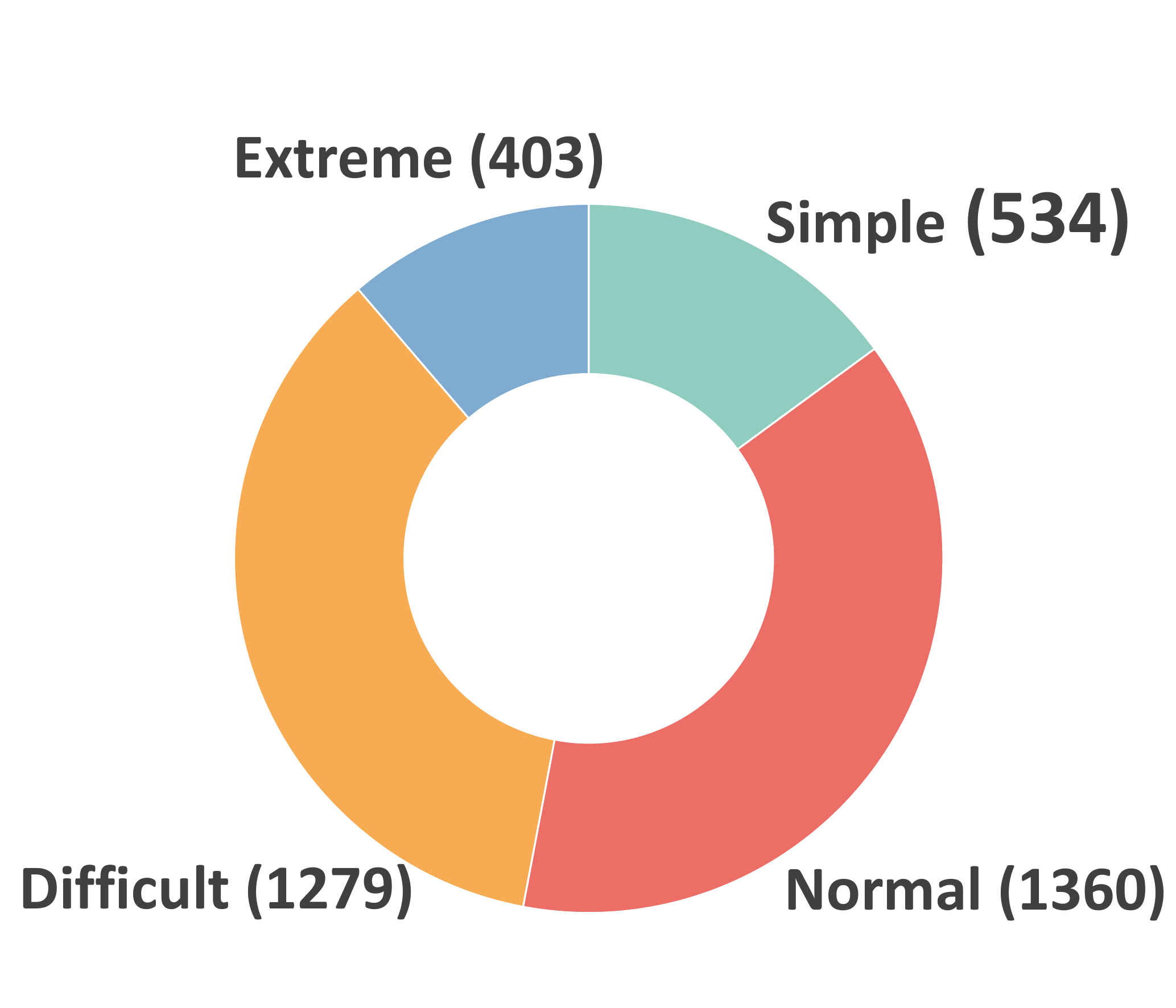}}
    \caption{Statistics on the number of ToF and MC questions at different difficulty levels.}
    \label{figure3}
\end{figure}

\paragraph{Objective Questions}
For ToF and MC, we use \emph{accuracy} as a basic performance measure for each question.
Then, we design a scoring scheme based on \emph{difficulty coefficient} to assess the overall performance of each LLM.
We calculate the difficulty coefficient $Dc_i$ of each question $i$ based on the accuracy of all LLMs' answers.
Specifically, we have $Dc_{i} = 1 - \frac{\text{num}^{\text{acc}}_i}{\text{num}^{\text{llm}}_i}$, where $\text{num}^{\text{acc}}_i$ is the number of LLMs that correctly answer question $i$ and $\text{num}^{\text{llm}}_i$ is the total number of LLMs in the evaluation.
All questions are then divided into four \emph{difficulty levels}, namely \emph{simple}, \emph{normal}, \emph{difficult}, and \emph{extreme}, with different weights $w_i$ based on $Dc_i$ in the scoring scheme as follows:
\begin{equation*}
    w_i = \begin{cases}
            0.5, & Dc_i \in [0,0.2) \\
            1, & Dc_i \in [0.2,0.5) \\
            1.5, & Dc_i \in [0.5,0.8) \\
            2, & Dc_i \in [0.8,1].
          \end{cases}
\end{equation*}
Figure~\ref{figure3} presents the number of questions at each difficulty level in ToF and MC.
For PA, we assign a weight of $w_i = 3$ to each question $i$ and use the following scoring rules: (1) A completely correct answer gets $3$ points; (2) a partially correct answer gets $1$ point; and (3) a completely incorrect answer does not get any points.

\paragraph{Subjective Questions}
ES and CA are open-ended questions with no unique answers.
CA is even harder than ES because it may require the integration of multiple pieces of knowledge to answer, which requires more strong reasoning ability.
Therefore, we assign a weight of $5$ to each ES question and $10$ to each CA question.
Human evaluations are often necessary to accurately assess the quality of answers to these questions but incur huge labor costs.
Alternatively, we introduce an automated scoring scheme based on LLMs \citep{gaokao, LME}.
We set up prompts for GPT-4o to act as a referee to rate the responses of all other LLMs.
Furthermore, we meticulously assign a score to every sub-question within a CA question, enabling the referee to furnish precise rationales for their grading decisions. We also manually score the responses of some LLMs for CA questions and compute the Spearman and Kendall correlation coefficients to quantify the agreement between the human and GPT-4o scoring results in Table~\ref{table-correlation coefficients}, which indicates that the scores provided by GPT-4o are mostly consistent with those of human evaluations.

Finally, the total score of an LLM for each type of question is the sum of the scores of all (partially) correctly answered questions.
We follow the common practice of standardized medical examinations to assign a fixed proportion to each type of question (10 for ToF, 40 for MC, 10 for PA, 30 for ES, and 10 for CA).
The integrated score of an LLM is thus calculated as the weighted sum of the total score for each question type, i.e., $\mathrm{S}_{\text{total}} = \mathrm{S}_{\text{ToF}} \times {10} + \mathrm{S}_{\text{MC}} \times {40} + \mathrm{S}_{\text{PA}} \times{10} + \mathrm{S}_{\text{ES}} \times{30} + \mathrm{S}_{\text{CA}} \times{10}$.
\begin{table}[t]
    \small
    \setlength{\tabcolsep}{3.5pt}
    \centering
    \begin{tabular}{ccccc}
        \toprule
        \multirow{2}{*}{\textbf{Model}} & \multicolumn{2}{c}{\textbf{Score}} & \multirow{2}{*}{\textbf{Spearman}} & \multirow{2}{*}{\textbf{Kendall}} \\
        \cmidrule{2-3} & Human & GPT-4o \\
        \midrule
        BianQue-7B   & 139   & 150   & 0.7508 & 0.6430 \\
        ChatGLM3-6B  & 296   & 231   & 0.7328 & 0.6325 \\
        Qwen1.5-72B  & 453.5 & 425   & 0.8079 & 0.7149 \\
        GPT3.5-turbo & 315   & 300.5 & 0.7732 & 0.6511 \\
        \bottomrule
    \end{tabular}
    \caption{Correlation between human and GPT-4o evaluations for CA questions.}
    \label{table-correlation coefficients}
\end{table}

\begin{table*}[t]
    \small
    \centering
    \begin{tabular}{ccccccccccccc}
        \toprule
        \textbf{Domain} & \textbf{Model} & \textbf{ToF} & \textbf{MC} & \textbf{PA} & \textbf{ES} & \textbf{CA} & $\mathrm{S}_{\text{total}}$  \\
        \midrule
        \multirow{4}{*}{\textbf{Medical LLM}}  &	BianQue (7B)	&	0	&	0	&	0	&	13.01	&	2.24	&	15.25	\\
        &	QiZhenGPT (13B)	&	0	&	0	&	0	&	15.88	&	2.54	&	18.41	\\
        &	PULSE (7B)	&	4.63	&	9.39	&	1.12	&	15.37	&	3.18	&	33.69	\\
        &	PULSE (20B)	&	5.03	&	15.59	&	1.71	&	15.25	&	3.70	&	41.28	\\
        \midrule
        \multirow{12}{*}{\textbf{Open-Source LLM}}  &	Baichuan2-7B	&	5.22	&	15.09	&	1.27	&	16.77	&	4.25	&	42.59	\\
        &	Baichuan2-13B	&	5.27	&	16.28	&	1.55	&	18.46	&	4.66	&	46.22	\\
        &	ChatGLM3-6B	&	4.88	&	12.06	&	1.41	&	16.10	&	3.45	&	37.91	\\
        &	InternLM2-7B	&	5.12	&	16.38	&	2.21	&	20.16	&	5.24	&	49.11	\\
        &	InternLM2-20B	&	4.67	&	16.91	&	2.51	&	20.49	&	5.73	&	50.31	\\
        &	LLaMa3-8B	&	4.67	&	12.16	&	1.67	&	13.67	&	2.21	&	34.37	\\
        &	LLaMa3-70B	&	6.53	&	22.05	&	4.76	&	17.79	&	4.28	&	55.41	\\
        &	Qwen1.5-7B	&	5.28	&	12.80	&	1.68	&	15.17	&	3.76	&	38.70	\\
        &	Qwen1.5-14B	&	6.37	&	20.13	&	4.43	&	16.62	&	4.87	&	52.41	\\
        &	Qwen-72B	&	6.78	&	30.79	&	5.65	&	19.03	&	6.34	&	68.60	\\
        &	Mixtral-8x7B	&	4.93	&	12.75	&	2.13	&	14.70	&	2.66	&	37.17	\\
        &	Mixtral-8x22B	&	5.92	&	15.39	&	3.95	&	16.73	&	3.73	&	45.71	\\
        \midrule
        \multirow{4}{*}{\textbf{Commercial LLM}}  &	GLM-4	&	6.98	&	28.12	&	5.39	&	22.63	&	6.61	&	69.74	\\
        &	ERNIE-3.5	&	7.67	&	27.11	&	7.68	&	21.74	&	7.67	&	71.87	\\
        &	Qwen-MAX	&	7.58	&	30.16	&	7.17	&	23.14	&	7.69	&	75.74	\\
        &	GPT3.5-turbo	&	5.47	&	15.11	&	3.82	&	17.50	&	4.49	&	46.38	\\
        \midrule
        \multicolumn{2}{c}{\textbf{Full Score}} &  10&  40&  10&  30&  10&  100\\
        \bottomrule
    \end{tabular}
    \caption{Scores of different LLMs for the five question types on the PediaBench dataset.}
    \label{table-main results}
\end{table*}

\section{Experiments}
\label{sec-exp}

\subsection{Experimental Setup}

We validate PediaBench through experiments with 20 general-purpose and medical LLMs, including open-source and commercial models of various scales.
Specifically, the models we use include: (1) medical LLMs such as BianQue \citep{bianque}, QiZhenGPT \cite{QiZhenGPT}, PULSE-7B, and PULSE-20B \citep{pulse};
(2) the GLM series \citep{GLM}, including ChatGLM3-6B and GLM4;
(3) the Baichuan series \citep{Baichuan2}, comprising Baichuan2-7B and Baichuan2-13B;
(4) the Qwen series \citep{qwen}, including Qwen1.5-7B, Qwen1.5-14B, Qwen1.5-72B, and Qwen-MAX;
(5) the InternLM series \citep{Internlm2}, with InternLM2-7B and InternLM2-20B;
(6) the LLaMA3 series \citep{LLaMA}, featuring LLaMA3-8B and LLaMA3-70B;
(7) the Sparse Mixture of Experts language models, namely Mixtral 8x7B and Mixtral 8x22B \citep{mixtral};
(8) GPT3.5-turbo;
and (9) ERNIE-3.5 (We choose ERNIE-3.5-8K-0329) \citep{ernie}.
We deploy and evaluate open-source LLMs locally using pretrained weights on a server with eight NVIDIA RTX A6000 GPUs.
We access and evaluate commercial LLMs through their official APIs.
More detailed information on these LLMs is given in Appendix~\ref{app-llms}.

We developed a standardized set of prompts for all LLMs.
This ensures that each LLM can generate the desired responses to each type of question based on the pre-specified requirements, thereby achieving a standardized and fair evaluation process.
Appendix~\ref{app:prompts} details the prompts for different types of questions.
We adopted the zero-shot prompt setting in all experiments.

\subsection{Evaluation Results}

The results for the overall performance of LLMs are shown in Table~\ref{table-main results}.
We present the measures of each LLM for the five types of questions.
Note that GPT-4o can achieve the highest scores for three types of objective questions but is omitted from Table~\ref{table-main results} to avoid unfair comparison since GPT-4o serves as a referee to assess subjective questions.
BianQue-7B and QiZhenGPT-13B cannot correctly understand and follow the instructions for the objective questions, and thus, their scores are $0$ for these types.
In terms of ToF, PULSE-7B obtains the lowest score of 4.63 among all models, which is not even half of the total score for ToF.
ERNIE-3.5 generally has the best performance for ToF and PA.
Generally, larger models achieve better performance than smaller ones in both objective and subjective questions.
However, GPT3.5-turbo exhibited a notably subpar performance, obtaining only half the full score.
Such results are quite surprising considering the strong capacities of GPT3.5-turbo in general and medical linguistic tasks.
We conjecture that this is because the GPT3.5-turbo training data lacks language resources in Chinese.

\begin{figure*}[t]
    \centering
    \subfigcapskip=0pt
    \subfigure[Scores (scaled to 0--100) of open-source LLMs for five question types]{
        \includegraphics[width=0.47\linewidth]{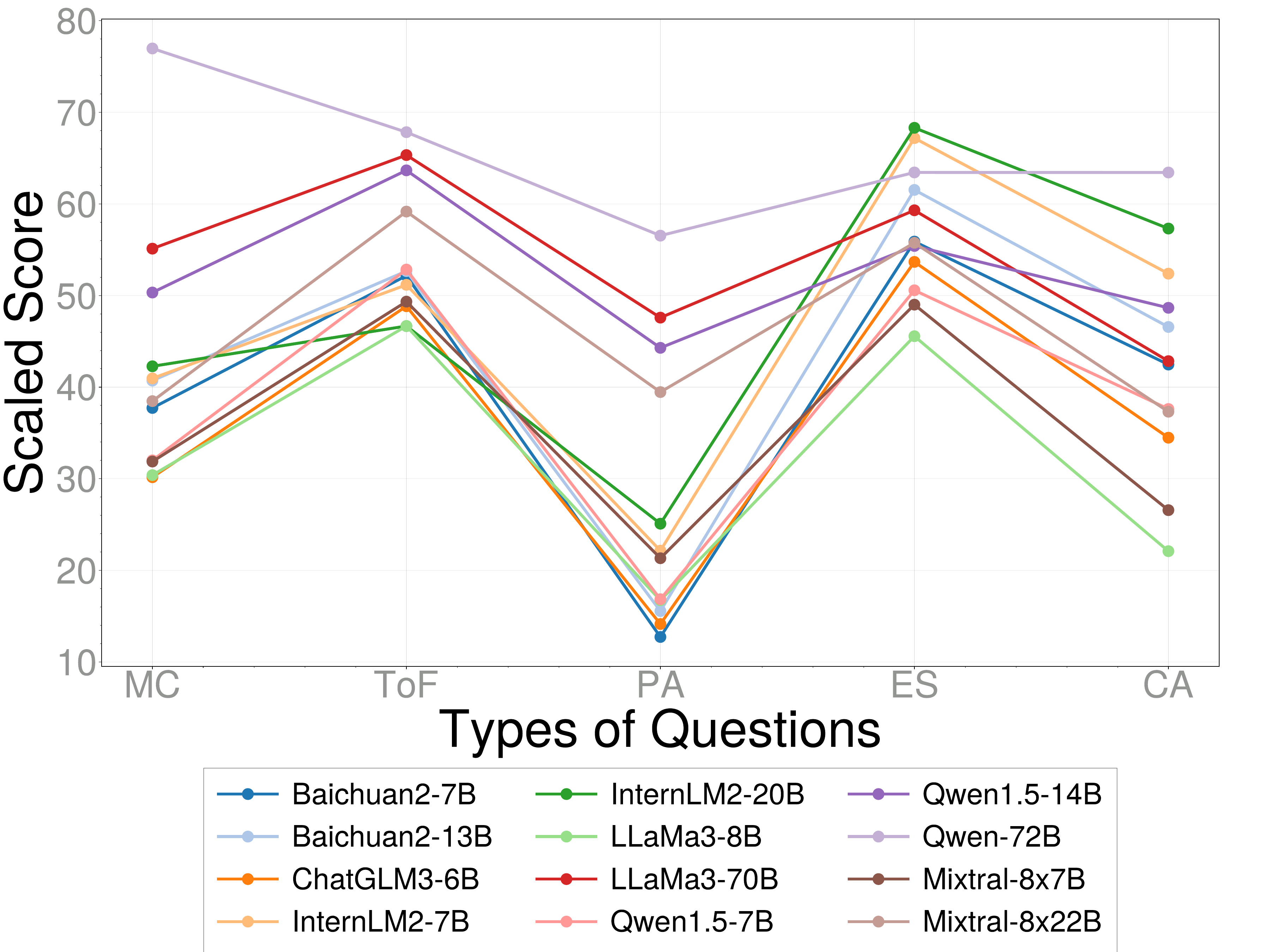}
        \label{figure4a}
    }
    \hspace{1em}
    \subfigure[Four-dimensional scores of LLMs by GPT-4o for ES]{
        \includegraphics[width=0.46\linewidth]{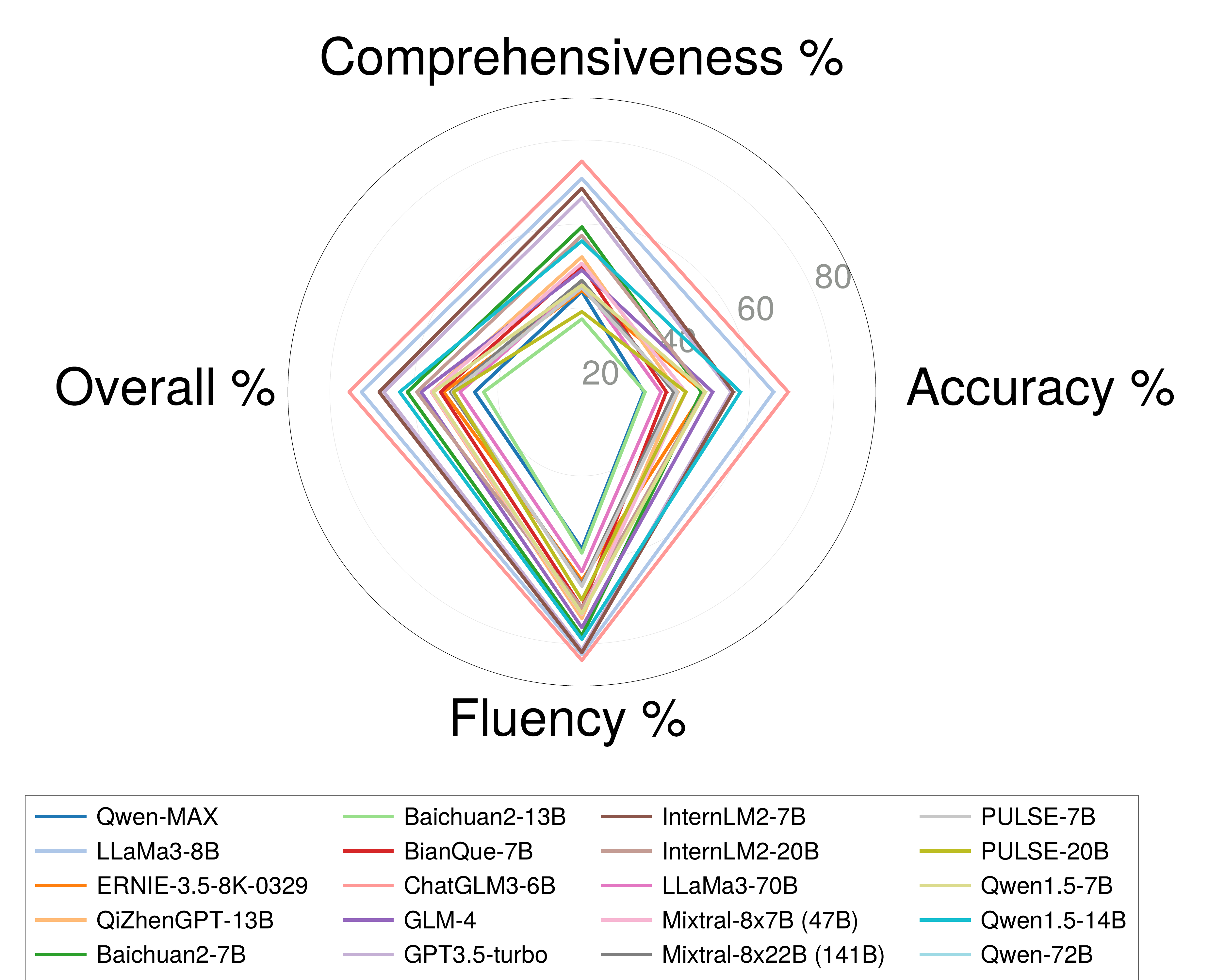}
        \label{figure4b}
    }
    \caption{Illustrations of evaluation results for open-source LLMs.}
    \label{figure4}
\end{figure*}

Figure~\ref{figure4a} displays the scores (scaled in the range of $[0, 100]$) of open-source LLMs for the five types of questions.
We observe that most LLMs obtain higher accuracy rates in ToF and ES questions.
In particular, PA questions are the most discriminatory type, where most LLMs exhibit poor scores.
This confirms the effectiveness of introducing PA into the PediaBench dataset, as this type of question is not included in most existing benchmarks.
We also find that CA generally yields lower scores than ES.
This is not surprising because CA requires the LLM to perform diagnoses and requires more precise and nuanced responses.
In addition, Figure~\ref{figure4b} shows the scores provided by GPT-4o for open-source LLMs in ES questions in four dimensions: \emph{accuracy}, \emph{comprehensiveness}, \emph{fluency}, and \emph{overall evaluation}.
Most LLMs perform well in terms of fluency but require further improvements in terms of comprehensiveness and accuracy.
Especially in the medical field, \emph{accuracy} should be considered a crucial and indispensable issue.
Therefore, in-depth investigations should be performed to avoid hallucination and ensure the correctness of the LLM responses.

In addition, we observe that for some types of questions (especially subjective questions), the LLaMA and Mixtral model series often provide answers that are mixtures of Chinese and English.
Although we introduce additional specifications in the prompts, it seems that this issue is not fully addressed.
Moreover, we also find that LLMs are not entirely clear about certain details, resulting in factual errors in their responses.
One example is that, for questions regarding the preferred treatment drugs for diseases, LLMs are capable of providing a general classification of drugs but often experience confusion when describing specific drugs in detail.
As shown in Appendix \ref{examples}, while PULSE-7B understands that the preferred drug treatment for the myasthenia crisis in children with myasthenia gravies is acetylcholinesterase inhibitor, it mistakenly refers to the more specific drug name such as pyridostigmine bromide as neostigmine.

\paragraph{Summary}
Our main experimental findings are summarized as follows:
\begin{itemize}
    \item Until now, only several commercial LLMs and Qwen-72B have achieved passing overall scores (i.e., $\geq 60$). Qwen-MAX achieved the highest total score of 75.74, firmly securing its top position but still far from achieving excellence in the exam (note that a total score of GPT-4o is nearly 80, which is not presented due to a potentially unfair comparison). Medical LLMs received lower scores due to poor instruction-following, reasoning, and generation capacity.
    On the one hand, this confirms that PediaBench is challenging and discriminative enough to assess the applicability of LLMs in pediatric scenarios.
    On the other hand, this also indicates that, despite the commendable efforts, there remains a huge gap to meet the requirement for deploying LLMs as AI assistants for pediatricians.
    \item Small LLMs can sometimes show good performance in the PediaBench dataset. For example, Qwen1.5-14B, with only 14B parameters, outperforms GPT3.5-turbo and Mixtral-8x22B with many more parameters. This could potentially be explained by the fact that GPT3.5-turbo and Mixtral-8x22B may lack sufficient exposure to Chinese medical corpora during their training. However, large-scale and commercial models exhibit significantly better performance in most cases.
    \item For future research and practice of medical LLMs, we have the following suggestions. First, the injection of medical knowledge \cite{Qilin-Med} is still an effective way to improve the QA capability of LLMs. By analyzing the examples in Appendix \ref{examples}, we find that many factual errors LLMs make come from misunderstanding in medical knowledge and thus cannot be resolved without knowledge injection.
    Second, some named entities (e.g., diseases and drugs) cannot be recognized by LLMs. Retrieval Augmented Generation (RAG) \cite{wang2024augmentingblackboxllmsmedical} may be a potential solution to this problem.
    Third, for CA questions, LLMs still suffer from low reasoning capacity, and Chain-of-Thought (CoT) prompting \cite{gramopadhye2024shotchainofthoughtdrivenreasoning} is required to improve their performance.
\end{itemize}

\section{Conclusion}

In this paper, we introduce PediaBench, a comprehensive Chinese benchmark dataset encompassing 12 pediatric disease groups to evaluate the QA capacity of LLMs.
Specifically, PediaBench consists of 4,117 objective and 1,632 subjective questions.
It adopts an integrated scoring criterion based on different difficulty levels to thoroughly assess the proficiency of an LLM in pediatric problems.
Finally, the effectiveness of PediaBench is confirmed through extensive experiments on 20 open-source and commercial LLMs.
Through an in-depth analysis of experimental results, we offer insights into the ability of LLMs for pediatric QA in Chinese, highlighting their limitations for improvement.

\section*{Limitations}
Despite the abundance of pediatric questions in the PediaBench dataset, it still cannot encompass many pediatric diseases and their corresponding treatments in the real world.
Therefore, the PediaBench dataset should be maintained with a continual effort for better coverage.
Currently, PediaBench is focusing primarily on pediatrics. In future work, we plan to extend it to more medical departments.
In addition, although the scoring method based on GPT-4o strongly correlates with human scoring, we cannot fully determine whether GPT-4o scoring tends towards a certain style, leading to potential biases. To address this issue, we will consider more rigorous scoring strategies in the future, such as using multiple LLMs for ensembling \cite{LME}.

\section*{Ethics Statement}
All data sources we use to construct the PediaBench dataset are publicly available and free to use without copyright infringement.
All questions in the PediaBench dataset have been appropriately anonymized so that they do not contain sensitive private information about patients.
We do not foresee any other possible negative societal impacts of this work.

\bibliography{ref} 

\appendix

\begin{table*}[!ht]
    \centering
    \small
    \begin{tabular}{l p{10.6cm}}
        \toprule
        \textbf{Question Type} & \textbf{Description}\\
        \midrule
        A1 (Single-Select Multiple-Choice) & The question consists of a single-sentence stem and five alternative answers, among which only one is the best answer.\\
        \midrule
        A2 (Individual Case Analysis) & The question features a brief, small instance as the stem and a question with five alternative answers, among which the examinee should select the only best answer.\\
        \midrule
        A3/A4 (Multiple Case Analysis)  & The question features a brief, small instance as the stem and multiple questions, each with five alternative answers, among which the examinee should select the only best answer for each question.\\
        \midrule
        B1 (Shared-Answer) & Given the same five alternative answers and a set of questions, the examinee should select the best answer for each question.\\
        \midrule
        True or False & Given a sentence or a paragraph, determine whether the statement is accurate.\\
        \midrule
        Essay/Short-Answer & Provide a concise answer to the question, elaborating on the issue.\\
        \midrule
        Case Analysis & Present one or more questions based on a case study of a simulated clinical scenario, requiring the examinee to provide the corresponding answers according to the requirements.\\
        \bottomrule
    \end{tabular}
    \caption{Types of questions in standardized medical examinations.\label{type of question}}
\end{table*}
\begin{table*}[!ht]
    \small
    \centering
    \begin{tabular}{cccccc}
        \toprule
        \textbf{Name} & \textbf{\#Parameters} & \textbf{Context Window} & \textbf{Domain} & \textbf{Open Source} & \textbf{How Accessed} \\
        \midrule
        BianQue & 7B & / & Medical & Yes & Weights\\
        QiZhenGPT & 13B & / & Medical & Yes & Weights\\
        PULSE-7B & 7B & / & Medical & Yes & Weights\\
        PULSE-20B & 20B & / & Medical & Yes & Weights\\
        \midrule
        ChatGLM3-6B & 6B & 8K & General & Yes & Weights\\
        Baichuan2-7B-Chat & 7B & 4K & General & Yes & Weights\\
        Qwen1.5-7B & 7B & 32K & General & Yes & Weights\\
        InternLM2-7B & 7B & 200K & General & Yes & Weights\\
        Baichuan2-13B-Chat & 13B & 4K & General & Yes & Weights\\
        Qwen-14B-Chat & 14B & 32K & General & Yes & Weights\\
        Qwen-72B-Chat & 72B & 32K & General & Yes & Weights\\
        \midrule
        Mixtral 8x7B & 47B & 32K & General & Yes & API\\
        InternLM2-20B & 20B & 200K & General & Yes & API\\
        LLaMa3-70B & 70B & 8K & General & Yes & API\\
        LLaMa3-8B & 8B & 8K & General & Yes & API\\
        Mixtral 8x22B & 141B & 32K & General & Yes & API\\
        \midrule
        Qwen-MAX & n/a & 8K & General & No & API\\
        GLM-4 & n/a & 128K & General & No & API\\
        ERNIE-3.5 & n/a & 8K & General & No & API\\
        ChatGPT3.5-turbo & n/a & 16K & General & No & API\\
        GPT-4o (*examiner) & n/a & 128K & General & No & API\\
        \bottomrule
    \end{tabular}
    \caption{List of LLMs evaluated and used in the experiments.}
    \label{model}
\end{table*}

\section{Detailed Description of Question Types}

Table~\ref{type of question} summarizes the different questions in standardized medical examinations.
PediaBench covers all these questions and involves pairing questions not included in existing medical examinations and benchmarks.

\section{List of LLMs in the Experiments}
\label{app-llms}

Table~\ref{model} presents detailed information on the LLMs we evaluate and use, where ``Domain'' indicates whether an LLM is of general purpose or specialized in the medical domain, ``\#Parameters''  presents the number of parameters of an LLM (``n/a'' for commercial models with disclosed parameter numbers), ``Context Window'' reveals the size of the context window of an LLM, ``How Accessed'' indicates how we accessed an LLM for experimentation (open-source models are obtained through their weights and deployed locally on our servers and commercial models are accessed via their official APIs).

\section{Prompts}
\label{app:prompts}

\subsection{Prompt for Disease Group Classification}

\begin{figure*}[t]
    \centering
    \includegraphics[width=\linewidth]{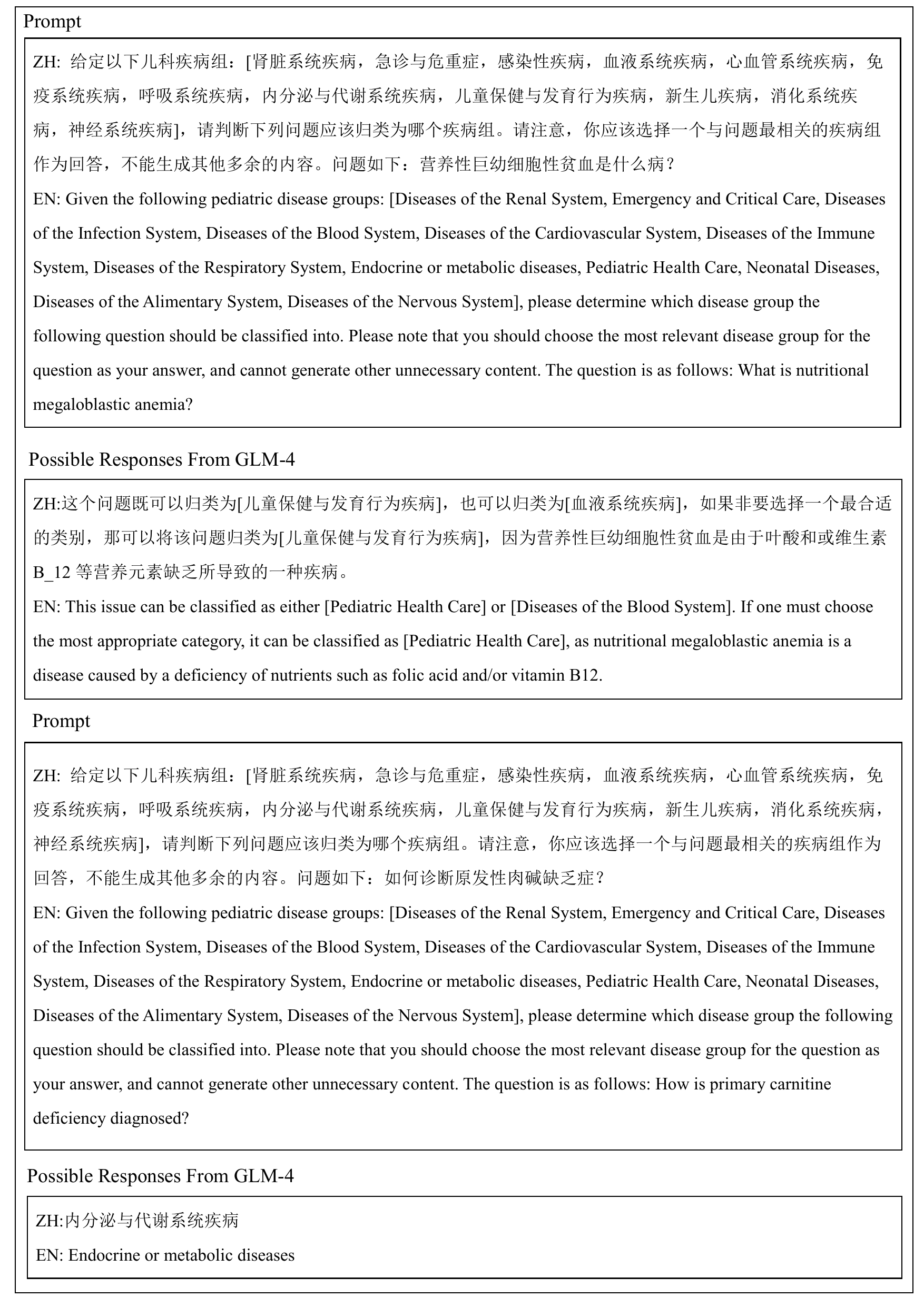}
    \caption{Prompt for disease group classification and an exemplar response of GLM-4.}
    \label{prompt-disease}
\end{figure*}

In this section, we present the prompt to classify the disease group of questions in PediaBench using GLM-4.
Specifically, we list the 12 disease groups and extract questions that have not yet been categorized.
Next, we set restrictive conditions to ensure that GLM-4 only returns the disease group that matches a question without irrelevant information.
Finally, we conduct a thorough review of the results labeled with GLM-4 and manually correct the questions that are misclassified.
Figure~\ref{prompt-disease} is an example of the prompt for disease group classification and the response of GLM-4.

\subsection{Prompt Templates for LLM Evaluation}

\begin{figure*}[t]
    \centering
    \includegraphics[width=\linewidth]{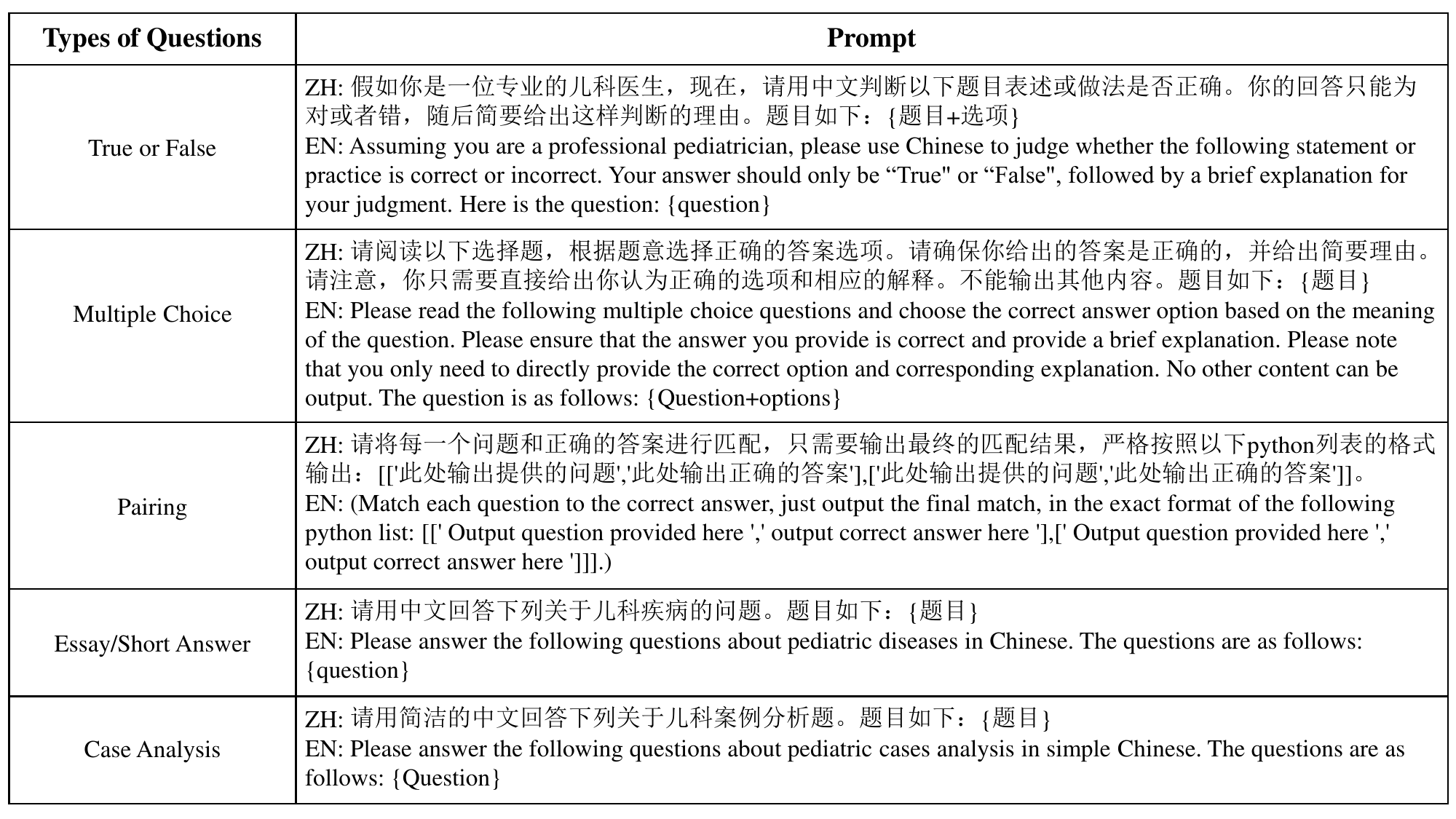}
    \caption{Prompt templates for different question types.}
    \label{figure_prompt}
\end{figure*}

Figure~\ref{figure_prompt} illustrates the templates of prompts for different question types when evaluating the LLMs in the experiments.

\subsection{Prompt for LLM as an Examiner}

\begin{figure*}[t]
    \centering
    \includegraphics[width=\linewidth]{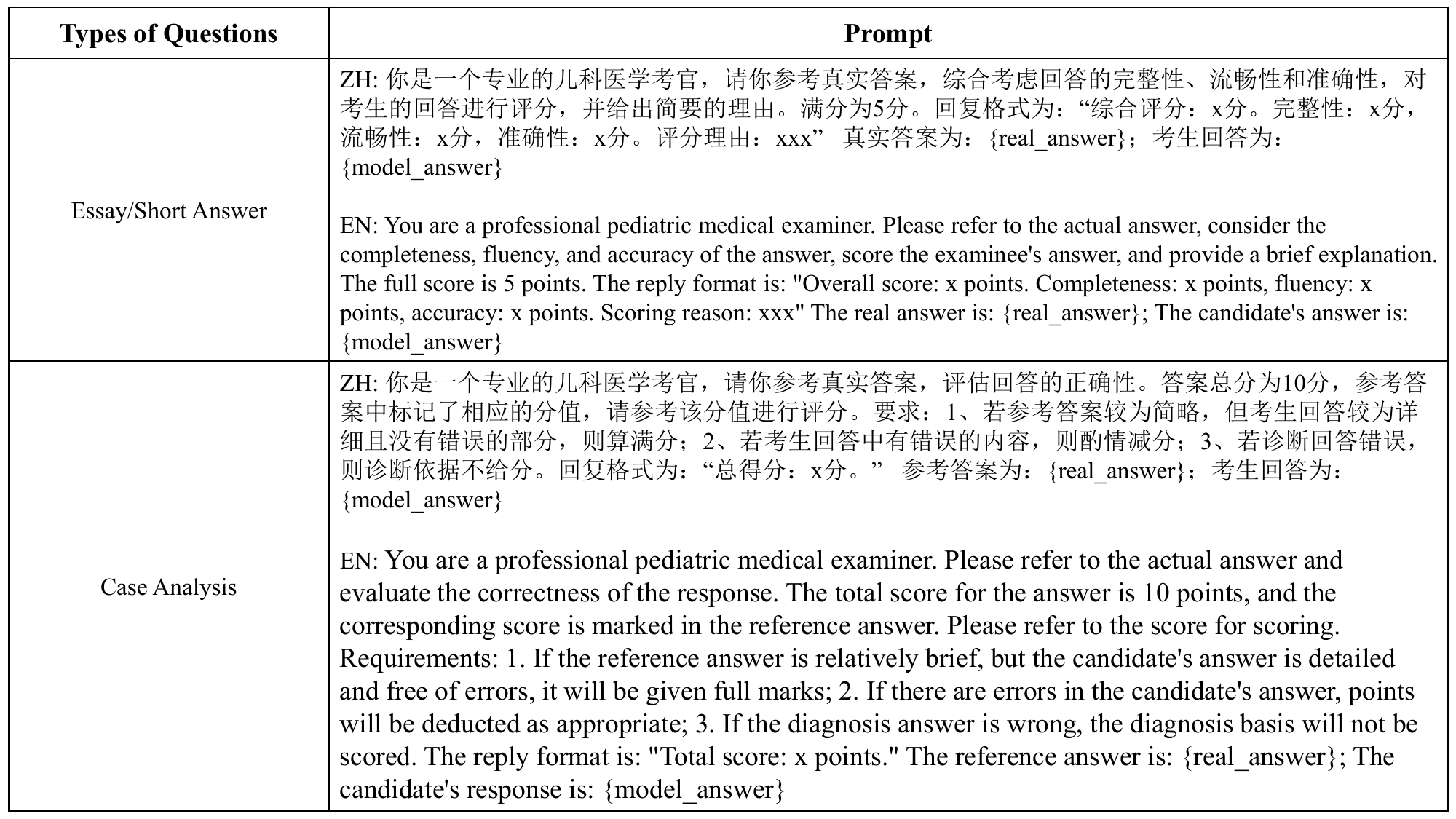}
    \caption{Prompts for GPT-4o as an examiner for scoring.}
    \label{examiner_prompt}
\end{figure*}

Figure~\ref{examiner_prompt} illustrates the prompts when using GPT-4o as an examiner. For essay/short answer questions, we instruct GPT-4o to grade responses based on four dimensions: \emph{fluency}, \emph{completeness}, \emph{accuracy}, and \emph{overall score}.
The scores are output in a specified format. For case analysis questions, we assign specific scores to each sub-question, instructing GPT to carefully grade according to these score standards, and finally aggregate the scores to obtain the total score for the entire question's response.


\section{Experimental Results for Different Disease Groups}

Table~\ref{table5} shows the scores of each LLM specifically for objective questions and ES questions across 12 disease groups. In order to quantify the score, we calculated the proportion of LLMs' score based on the inconsistent number of questions and scores for each disease group. Due to the inadequate instruction-following capabilities of BianQue (7B) and QiZhenGPT (13B), they struggled to provide precise responses to objective questions, resulting in surprisingly low scores. Intriguingly, despite Mixtral-8x7B's commendable instruction-following abilities, its error rate in providing answers remained remarkably high, ultimately leading to a comparable score with that of BianQue (7B) and QiZhenGPT (13B). We observe that most models achieve their highest scores in the two disease groups of HCDA of DImS.
Questions related to HCDA often involve basic medical knowledge, which does not require in-depth specialization. Moreover, many diseases in the DImS group share similar clinical manifestations and treatment strategies, providing models with clearer clues for answers.
No models can perform well for subjective questions in all performance measures across different groups of diseases. 

\begin{table*}[t]
    \centering
    \small
    \setlength{\tabcolsep}{4pt}
    \begin{tabular}{ccccccccccccc}
        \toprule
        \textbf{Model}  & \textbf{DInS} & \textbf{DRpS} & \textbf{HCDA} & \textbf{ECC} & \textbf{DImS} & \textbf{EMD} & \textbf{DNS} & \textbf{DRnS} & \textbf{DAS} & \textbf{DCS} & \textbf{DBS} & \textbf{ND} \\
        \midrule
        BianQue-7B  &  40.17   &  40.43   &  37.66   &  32.90   &  39.23   &  36.85   &  38.53   &  39.52   &\textbf{41.87}&  37.66   &  39.76   &  40.28   \\
        QiZhenGPT-13B  &  40.55   &  40.34   &  38.24   &  39.03   &\textbf{40.68}&  38.16   &  37.93   &  38.17   &  38.01   &  37.79   &  37.66   &  38.31   \\
        PULSE-7B  &  46.41   &  44.19   &\textbf{47.59}&  41.13   &  45.65   &  44.03   &  41.41   &  45.31   &  46.58   &  42.07   &  44.05   &  45.32   \\
        PULSE-20B  &  44.39   &  44.64   &  47.02   &  33.68   &  42.28   &  43.35   &  37.38   &  45.57   &  40.14   &  39.96   &  40.92   &\textbf{47.16}\\
        \midrule
        Baichuan2-7B  &  36.15   &  36.99   &\textbf{38.12}&  31.97   &  33.99   &  34.85   &  32.30   &  35.28   &  32.74   &  33.96   &  34.06   &  35.66   \\
        Baichuan2-13B  &\textbf{41.32}&  41.05   &  41.04   &  33.70   &  38.58   &  37.94   &  35.88   &  38.78   &  36.61   &  36.95   &  37.14   &  39.01   \\
        ChatGLM3-6B  &  45.90   &  48.41   &\textbf{48.98}&  44.39   &  46.25   &  43.07   &  43.60   &  47.95   &  48.25   &  42.89   &  44.61   &  45.32   \\
        InternLM2-7B  &  57.38   &\textbf{58.61}&  57.48   &  53.15   &  58.54   &  53.80   &  53.64   &  55.63   &  56.48   &  53.30   &  54.76   &  56.48   \\
        InternLM2-20B  &  55.24   &  56.98   &  57.66   &  54.38   &\textbf{57.98}&  55.78   &  52.37   &  55.33   &  52.21   &  51.79   &  55.16   &  55.38   \\
        LLaMa3-8B  &  46.03   &  45.97   &  48.47   &  44.02   &  48.74   &  46.12   &  45.20   &  46.81   &  48.00   &  45.99   &\textbf{48.86}&  47.78   \\
        LLaMa3-70B  &  52.19   &  51.94   &  53.45   &  50.89   &  54.03   &  52.92   &  52.17   &  53.28   &  53.93   &  50.17   &\textbf{55.16}&  52.27   \\
        Qwen1.5-7B  &  44.93   &  45.89   &  44.40   &  35.80   &  44.38   &  41.09   &  38.43   &\textbf{47.55}&  43.36   &  37.71   &  41.47   &  42.38   \\
        Qwen1.5-14B  &  53.42   &\textbf{56.71}&  52.90   &  51.23   &  54.34   &  52.84   &  52.56   &  54.89   &  52.93   &  52.09   &  53.44   &  51.21   \\
        Qwen-72B  &  68.11   &  68.16   &  67.62   &  65.55   &  65.06   &  65.94   &\textbf{68.86}&  67.69   &  71.54   &  68.22   &  67.20   &  66.57   \\
        Mixtral-8x7B  &  40.55   &  43.16   &  47.08   &  43.34   &\textbf{48.46}&  42.86   &  41.11   &  43.74   &  45.62   &  38.27   &  47.78   &  43.88   \\
        Mixtral-8x22B   &  42.76   &  47.72   &  48.51   &  44.39   &\textbf{50.94}&  46.66   &  43.01   &  47.06   &  44.91   &  44.25   &  50.82   &  46.60   \\
        \midrule
        GLM-4  &\textbf{74.16}&  73.08   &  70.42   &  65.31   &  71.40   &  73.56   &  72.55   &  69.56   &  74.05   &  71.80   &  72.05   &  68.80   \\
        ERNIE-3.5  &  60.99   &  59.15   &\textbf{63.24}&  55.79   &  58.99   &  58.81   &  59.23   &  59.17   &  58.13   &  56.84   &  57.92   &  58.83   \\
        Qwen-MAX  &  76.35   &  78.08   &  76.72   &  73.56   &  76.24   &\textbf{79.27}&  75.39   &  73.45   &  77.20   &  75.65   &  76.39   &  75.54   \\
        GPT3.5-turbo  &  61.02   &  62.57   &  60.11   &  61.86   &  62.36   &  59.89   &  57.79   &  62.33   &\textbf{62.80}&  58.34   &  61.41   &  60.31   \\
        \bottomrule
    \end{tabular}
    \caption{Scores of different LLMs in different disease groups on PediaBench. Here, the highest score in each disease group is highlighted in bold.}
    \label{table5}
\end{table*}

\section{Examples of Model Responses}
\label{examples}

This section selects some representative questions and answers from different LLMs.
These cases illustrate that LLMs still suffer from apparent shortcomings in pediatric QA.

\subsection{True-or-False Question}

\begin{figure*}[t]
    \centering
    \includegraphics[width=\linewidth]{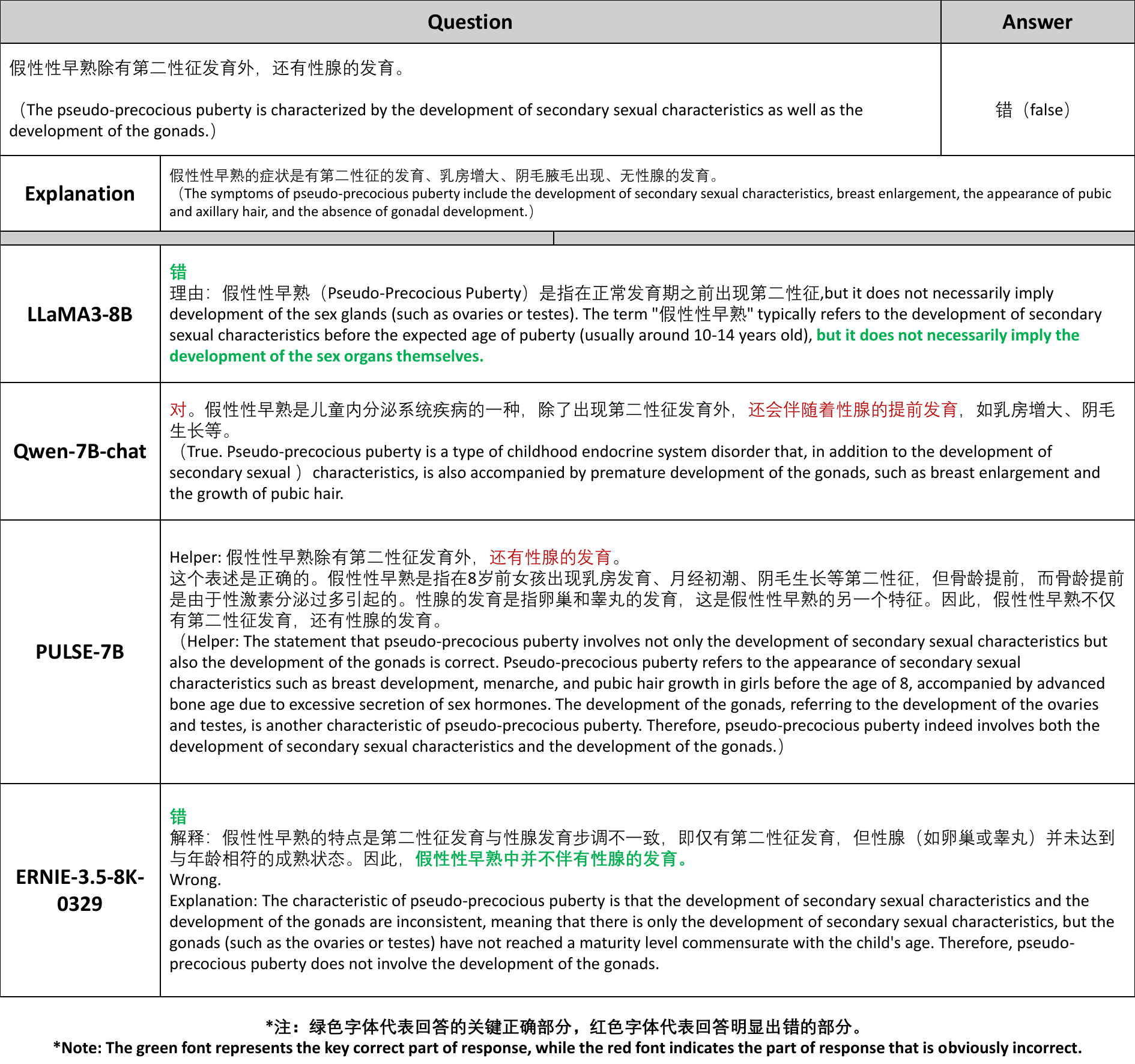}
    \caption{Answers and explanation notes of an exemplar true-or-false question.}
    \label{true or false}
\end{figure*}

For true-or-false questions, we compare the answers of PULSE-7B, LLaMA3-8B, Qwen-7B-Chat, and ERNIE-3.5, as shown in Figure~\ref{true or false}.
After in-depth analysis, we found an interesting phenomenon: models with smaller parameters, such as Qwen-7B-Chat, could judge statements correctly, even though there are errors in their explanations. Furthermore, despite our restrictions on the language generated by the models, the output of LLaMA3-8B always contains a mixture of Chinese and English.

\subsection{Multi-Choice Question}

\begin{figure*}[t]
    \centering
    \includegraphics[width=\linewidth]{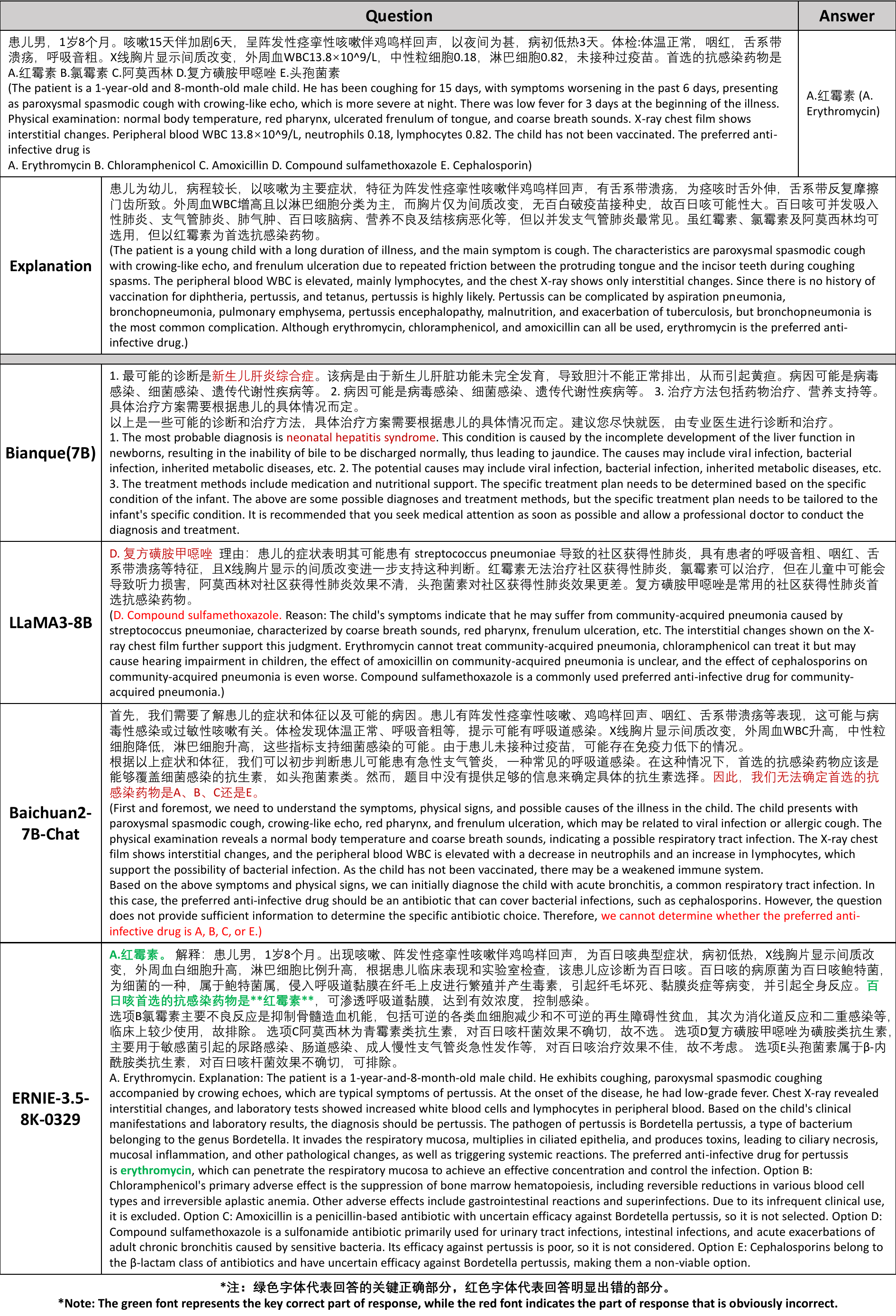}
    \caption{Answers and explanation notes of an exemplar multi-choice question.}
    \label{Multi-choice}
\end{figure*}

We conducted an in-depth analysis of a multiple-choice question on the disease of the infectious system.
The answers provided by several LLMs are shown in Figure ~\ref{Multi-choice}.
In this case, LLaMA3-8B and PULSE-20B provided incorrect options and explanations, failing to capture the key points of the question accurately.
Baichuan2-7B was even unable to make a reasonable choice.
However, ERNIE-3.5 not only selected the correct answer but also provided a detailed and reasonable explanation.

\subsection{Pairing Question}
\begin{figure*}[t]
    \centering
    \includegraphics[width=\linewidth]{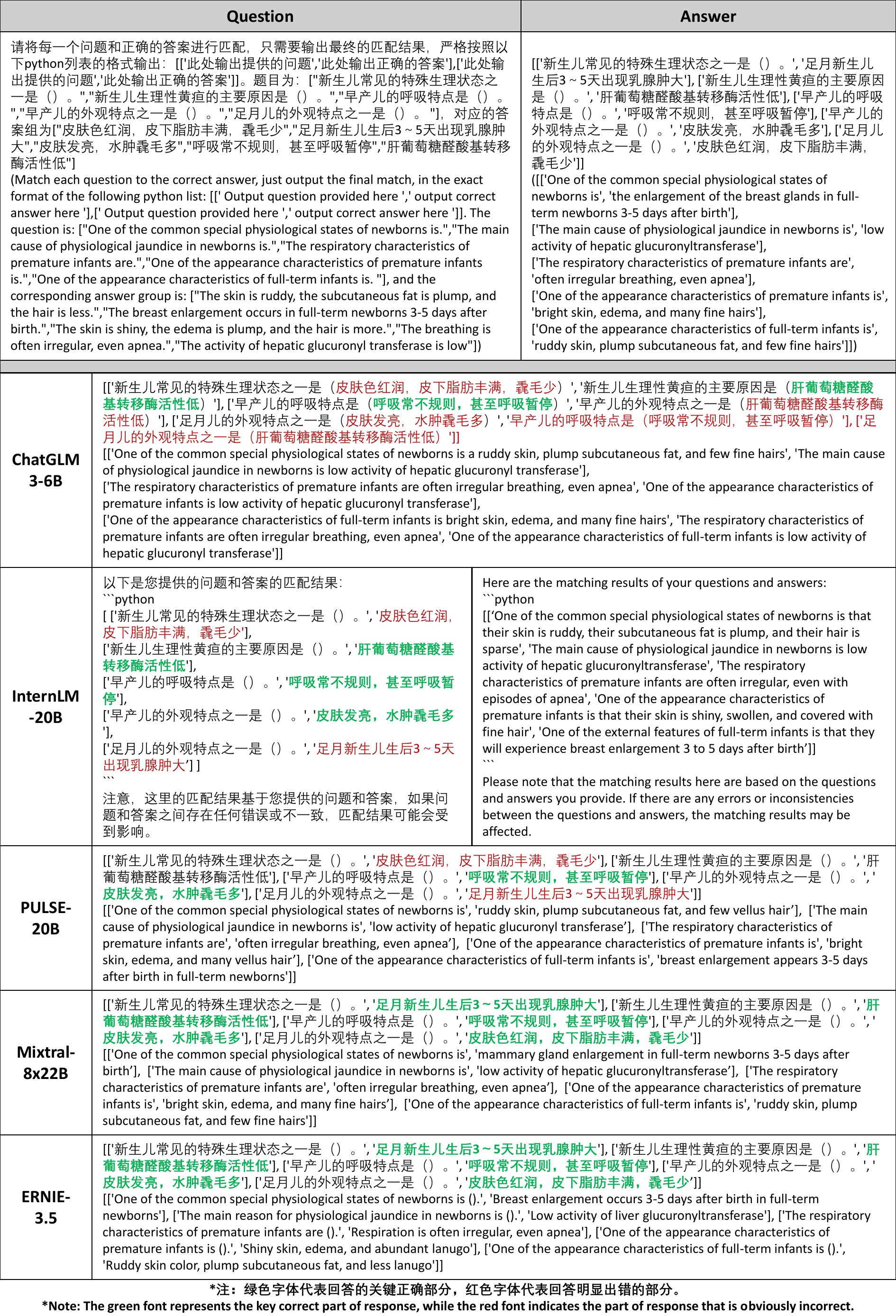}
    \caption{Answers and explanation notes of an exemplar pairing question.}
    \label{Pairing}
\end{figure*}

The answers to an exemplar pairing question are shown in Figure~\ref{Pairing}.
Only a few LLMs can provide complete and accurate results in this pairing question.
ChatGLM3-6B's answer contained duplicate answers, while InternLM-20B and Mixtral-8x22B provided incorrect matchings.
Unfortunately, PULSE-20B directly copied the order of the given answer sets in the question without performing any effective matching.
However, it should be mentioned that ERNIE-3.5 provided an entirely correct answer to this question, demonstrating its superior performance.

\subsection{Essay/Short-Answer Question}
\begin{figure*}[t]
    \centering
    \includegraphics[width=\linewidth]{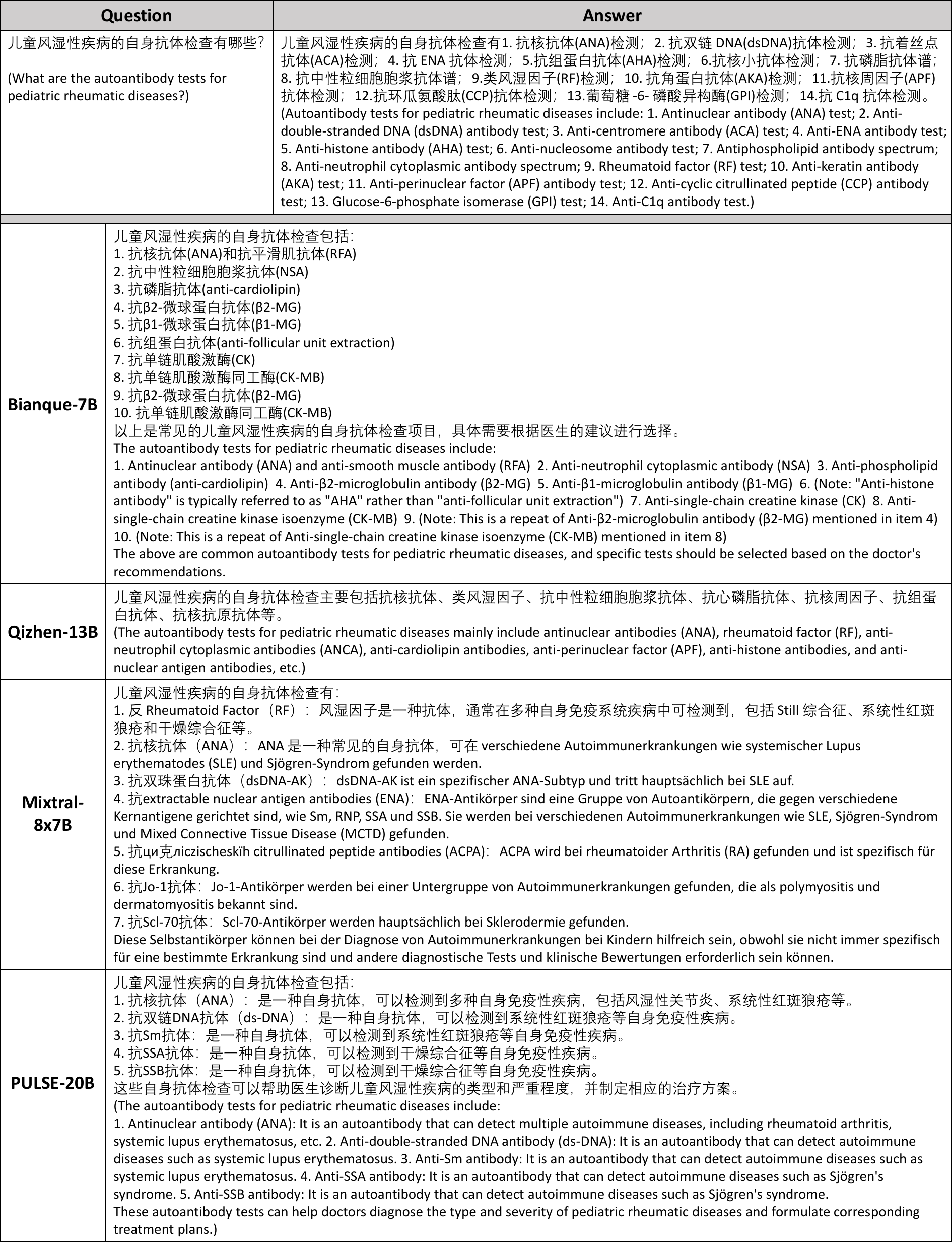}
    \caption{Answers and explanation notes of an exemplar essay/short-answer question.}
    \label{Essay-type Short Answer}
\end{figure*}

Figure~\ref{Essay-type Short Answer} presents several responses of LLMs to an essay/short-answer question.
Medical LLMs, such as BianQue and QiZhenGPT, tend to provide concise answers, while general-purpose LLMs prefer to elaborate and explain in more detail, resulting in richer content in their answers.
In particular, the response from PULSE-20B is not satisfactory, containing a significant amount of repetitive and logically unsound content, which, to some extent, affects the quality and precision of its response.

\subsection{Case Analysis Question}
\begin{figure*}[t]
    \centering
    \includegraphics[width=\linewidth]{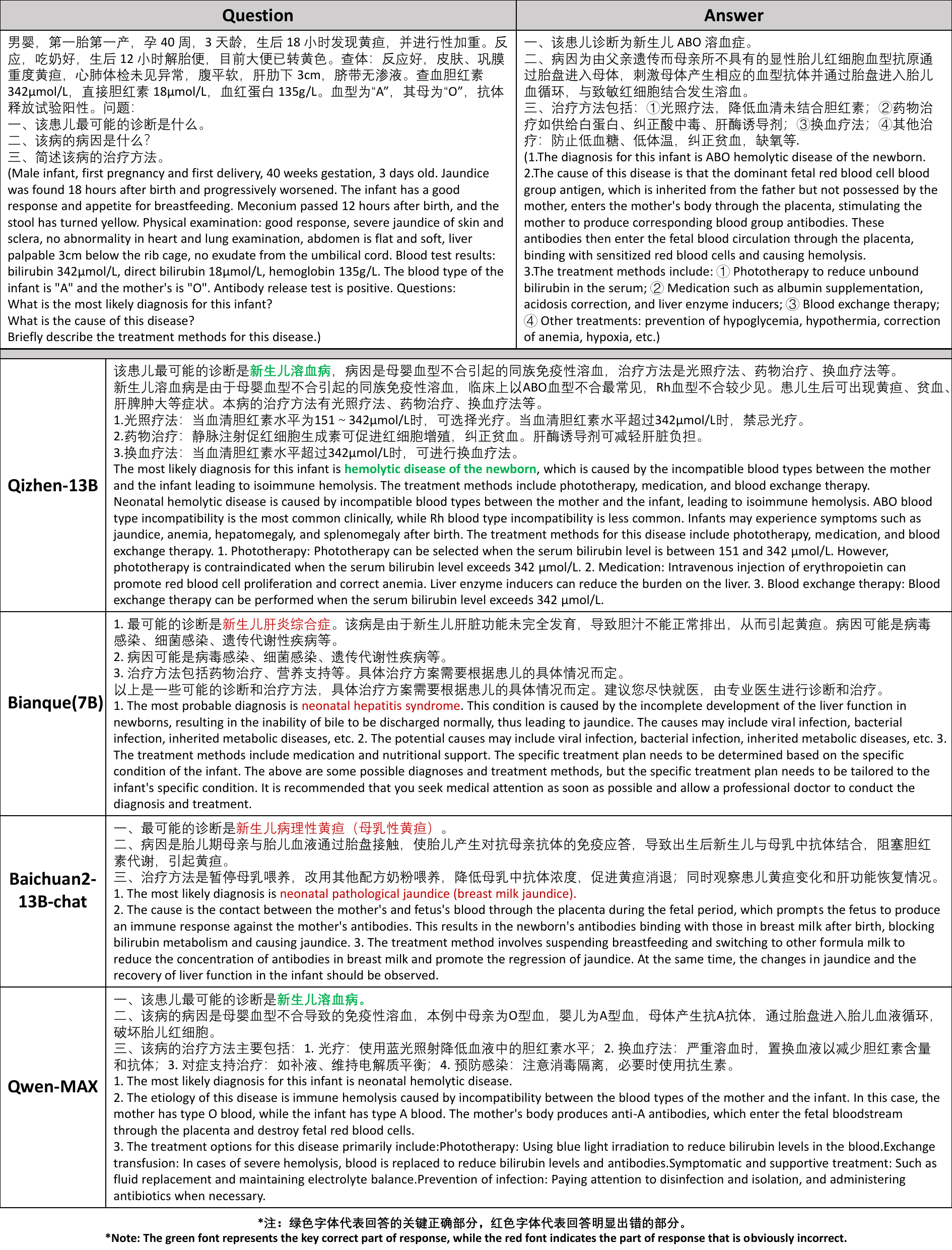}
    \caption{Answers and explanation notes of an exemplar case analysis question.}
    \label{Case Analysis}
\end{figure*}

In case analysis questions, the answer to each sub-question serves as the foundation for the solution to the subsequent sub-questions.
Therefore, once the answer to the previous question is incorrect, it will further mislead the conclusions of subsequent questions like dominoes.
In the example shown in Figure~\ref{Case Analysis}, their judgments and suggestions for subsequent questions become totally erroneous due to BianQue and Baichuan2-13B-Chat's misinterpretation of the first sub-question.

\end{document}